\documentclass[11pt,a4paper]{article}

\usepackage[hyperref]{acl2020}
\usepackage{times}
\usepackage{latexsym}

\usepackage{enumitem}
\usepackage{microtype}

\usepackage{graphicx} 
\usepackage{multirow}
\usepackage{latexsym}
\usepackage{amsmath}
\usepackage{amssymb}
\usepackage{mathrsfs}
\usepackage{enumerate}
\usepackage{bm}
\usepackage{courier}
\usepackage{color,xcolor,colortbl}
\usepackage{array}
\usepackage{color}
\usepackage[ruled,vlined]{algorithm2e}

\newcommand{\textcite}[1]{\citeauthor{#1} (\citeyear{#1})}

\aclfinalcopy 

\title{Learning to Customize Model Structures for \\Few-shot Dialogue Generation Tasks}

\author{Yiping Song$^1$, Zequn Liu$^1$, Wei Bi$^2$, Rui Yan$^3$, Ming Zhang$^1$\thanks{$^*$Corresponding author} \\
  $^1$Department of Computer Science, School of EECS, Peking University, Beijing, China \\
  $^2$Tencent AI Lab, Shenzhen, China \\
  $^3$Wangxuan Institute of Computer Technology, Peking University, Beijing, China\\
  \texttt{{songyiping,zequnliu,mzhang\_cs,ruiyan}@pku.edu.cn}\\
  \texttt{victoriabi@tencent.com }
  }

\date{}

\begin{document}
\maketitle
\begin{abstract}

Training the generative models with minimal corpus is one of the critical challenges for building open-domain dialogue systems. Existing methods tend to use the meta-learning framework which pre-trains the parameters on all non-target tasks then fine-tunes on the target task. However, fine-tuning distinguishes tasks from the parameter perspective but ignores the model-structure perspective, resulting in similar dialogue models for different tasks. In this paper, we propose an algorithm that can customize a unique dialogue model for each task in the few-shot setting.
In our approach, each dialogue model consists of a shared module, a gating module, and a private module. The first two modules are shared among all the tasks, while the third one will differentiate into different network structures to better capture the characteristics of the corresponding task. The extensive experiments on two datasets show that our method outperforms all the baselines in terms of task consistency, response quality, and diversity.

\end{abstract}

\section{Introduction}

Generative dialogue models often require a large amount
of dialogues for training, and it is challenging to build models that can adapt to new domains or tasks with limited data.
With recent advances in large-scale pre-training~\cite{elmo,ulmfit,gpt,bert}, 
we can first pre-train a generative model on large-scale dialogues from the non-target domains and then fine-tune on the task-specific data corpus~\cite{multi-relation-pre-train,pre-train-finetune,attention-pre-train}.
While pre-training is beneficial, such models still require sufficient task-specific data for fine-tuning.
They cannot achieve satisfying performance when very few examples are given~\cite{finetune-need-samples}.
Unfortunately, this is often the case in many dialogue generation scenarios.
For example, in personalized dialogue generation, we need to quickly adapt to the response style of a user's persona by just a few his or her dialogues~\cite{paml,persona}; in emotional dialogue generation, we need to generate a response catering to a new emoji using very few utterances containing this emoji~\cite{emotion-chat,mojitalk}. 
Hence, this is the focus of our paper - few-shot dialogue generation, i.e. training a generative model that can be generalized to a new task (domain) within $k$-shots of its dialogues. 
A few works have been proposed to consider few-shot dialogue generation as a meta-learning problem~\cite{paml,qian2019domain,mi2019meta}.
They all rely on the popular model-agnostic meta-learning (MAML) method~\cite{finn2017model}. 
Take building personalized dialogue models as an example, previous work treats learning dialogues with different personas as different tasks~\cite{paml,qian2019domain}.
They employ MAML to find an initialization of model parameters
by maximizing
the sensitivity of the loss function when applied
to new tasks.
For a target task, its dialogue model is obtained by fine-tuning the initial parameters from MAML with its task-specific training samples.


Despite the apparent success in few-shot dialogue generation, MAML still has limitations~\cite{context}. The goal of generative dialogue models is to build a function mapping a user query to its response, where the function is determined by both the model structure and parameters~\cite{automl}. By fine-tuning with a fixed model structure, MAML only searches the optimal parameter settings in the parameter optimization perspective but ignores the search of optimal network structures in the structure optimization perspective. 
Moreover, 
language data are inherently discrete and dialogue models are less vulnerable to input changes than image-related models~\cite{niu2018adversarial}, which means gradients calculated from a few sentences may not be enough to change the output word from one to another. 
Thus there is a need to develop an effective way to adjust MAML for large model diversity in dialogue generation tasks.
In this paper, we propose the Customized Model Agnostic Meta-Learning algorithm (CMAML) that is able to customize dialogue models in both parameter and model structure perspective under the MAML framework.
The dialogue model of each task consists of three parts: a shared module to learn the general language generation ability and common characteristics among tasks, a private module to model the unique characteristic of this task, and a gate to absorb information from both shared and private modules then generate the final outputs. The network structure and parameters of the shared and gating modules are shared among all tasks, while the private module starts from the same network but differentiates into different structures to capture the task-specific characteristics.




In summary, our contributions are as follows:
\begin{itemize}[wide=0\parindent,noitemsep]
\item We propose the CMAML algorithm that can customize dialogue models with different network structures for different tasks in the few-shot setting. 
The algorithm is general and well unified to adapt to various few-shot generation scenarios.
\item We propose a pruning algorithm that can adjust the network structure for better fitting the training data. We use this strategy to customize unique dialogue models for different tasks.
\item We investigate two crucial impact factors for meta-learning based methods, i.e., the quantity of training data and task similarity. We then describe the situations where the meta-learning can outperform other fine-tuning methods. 
\end{itemize}

\section{Related Work}
{\bf Few-shot Dialogue Generation.}
The past few years have seen increasing attention on building dialogue models in few-shot settings, such as personalized chatbots that can quickly adapt to each user's profile or knowledge background~\cite{persona,paml}, or that respond with a specified emotion~\cite{emotion-chat,mojitalk}. 
Early solutions are to use explicit~\cite{tian2017acl,persona,emotion-chat}  or implicit~\cite{lijiwei,mojitalk,emotion-chat} task descriptions, then introduce this information into the generative models.  
However, these methods require manually created task descriptions, which are not available in many practical cases.  

An alternative promising solution to building few-shot dialogue models is the meta-learning methods, especially MAML~\cite{finn2017model}. 
\textcite{paml} propose to regard learning with the dialogue corpus of each user as a task and endow the personalized dialogue models by fine-tuning the initialized parameters on the task-specific data. 
\textcite{qian2019domain} and \textcite{mi2019meta} treat the learning from each domain in multi-domain task-oriented dialogue generation as a task, and apply MAML in a similar way.
All these methods do not change the original MAML but directly apply it to their scenarios due to the model-agnostic property of MAML. Thus, task differentiation always counts on fine-tuning, which only searches the best model for each task at the parameter level but not the model structure level.

\noindent
{\bf Meta-learning.}
Meta-learning has achieved promising results in many NLP problems recently due to its fast adaptation ability on a new task using very few training data~\cite{classficaiton-metric-meta,wang-etal-2019-extracting,relation-maml,alt-etal-2019-fine}.
In general, there are three categories of meta-learning methods: metric-based methods~\cite{vinyals2016matching,snell2017prototypical,sung2018learning,relation-metric-meta} which encode the samples into an embedding space along with a learned distance metric and then apply a matching algorithm, model-based methods~\cite{santoro2016meta,relation-meta-opt} which depend on the model structure design such as an external memory storage to facilitate the learning process, and optimization-based methods~\cite{finn2017model,andrychowicz2016learning,query-generation-meta} which learn a good network initialization from which fine-tuning can converge to the optimal point for a new task with only a few examples. 
Methods belonging to the first two are proposed for classification, and those in the third category are model-agnostic. Therefore, it is intuitive to apply the optimization-based methods, in which MAML is most popular, for dialogue generation tasks.

\begin{figure*}[!t]
\includegraphics[width=\textwidth,height=4.5cm]{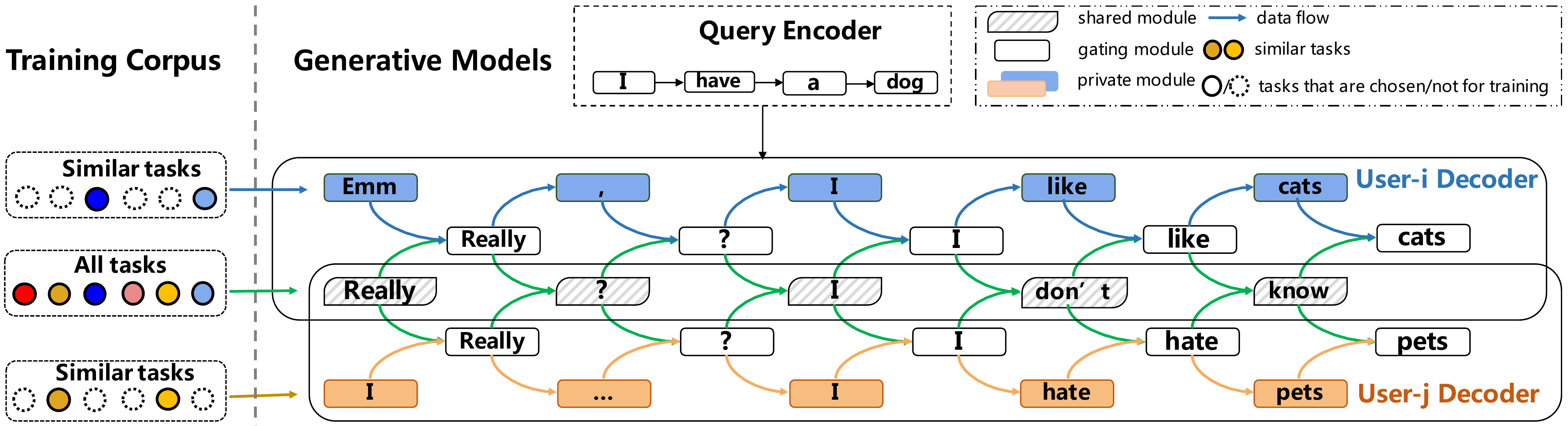}
\caption{The proposed CMAML algorithm applying on the personalized dialogue systems. Each customized dialogue model Seq2SPG consists of a shared, a private, and a gating module. The shared and gating module are the same among users and are trained on all tasks. The private module is unique for each user to describe this user's persona, and is trained on the corresponding and similar tasks. The lines in color indicate the data flow directions.}
\label{fig:over}
\end{figure*}

However, some researchers found that the original MAML has limited ability to model task-specific characteristics in the image or text classification scenarios~\cite{attentive,Sun_2019_CVPR,multi-task-few-shot}.
\textcite{attentive} build an attention layer over the convolutional layers, where the convolutional layer is for general features and the attention layer is for task-specific features. 
\textcite{Sun_2019_CVPR} propose to learn a task-specific shifting and scaling operation on the general shared feed-forward layers. However, the involved operations in these two methods such as shifting and scaling are designed for feed-forward networks, and can not be applied to the generative models which generally rely on Seq2seq~\cite{seq2seq} models with recurrent GRU~\cite{gru} or LSTM~\cite{lstm} cells. In this paper, we propose a new meta-learning algorithm based on MAML that can enhance task-specific characteristics for generation models. 


\section{Dialogue Model}
In this section, we firstly describe the network structure of the proposed dialogue model, and then briefly introduce its pre-training.

\subsection{Model Architecture}
\label{sec:dialogue model}
We aim to build dialogue models for different generation tasks in the few-shot setting. 
Now, we first describe the dialogue model of each task to be used in our training algorithm. It involves three network modules and noted as Seq2SPG (in Figure~\ref{fig:over}): 

\noindent \textbf{Shared Module.} It gains the basic ability to generate a sentence and thus its parameters are shared among all tasks. 
We employ a prevailing Seq2seq dialogue model~\cite{seq2seq-attention}. At each decoding step $t$, we feed the word $x_t$ and last hidden state $h_{t-1}$ to the decoding cell, and obtain an output distribution $o_s$ over the vocabulary.

\noindent \textbf{Private Module.} It aims at modeling the unique characteristics of each task. 
We design a multi-layer perception (MLP) in the decoder to fulfill this goal.
Each task has its unique MLP network,
which starts from the same initialization and then evolves into different structures during training. 
At each decoding step $t$, the MLP takes the word $x_t$ and the output $h_{t-1}$ of the shared module at step $t-1$ as input, then outputs a distribution $o_p$ over the vocabulary. In our experiments, we also explore different inputs for the private module.

\noindent \textbf{Gating Module.} 
We use a gate to fuse information from the shared and private modules: 
\begin{equation}
\begin{aligned}
\label{euq: gating}
g_s &= \tanh (W_s[o_s, o_p] + b_s) \\
g_p &= \tanh (W_p[o_s, o_p] + b_p) \\
o &= g_s \circ o_s + g_p \circ o_p
\end{aligned}
\end{equation}
where $W_s$, $W_p$, $b_s$, $b_p$ are parameters, $\circ$ is element-wise
product, and $o$ is the word distribution.

\subsection{Training Overview}
 

For the rest of the paper, $p(\mathcal{T})$ denotes the task distribution, $T_i$ denotes the $i$-th task to be trained, $D_i^{train}$ and $D_i^{valid}$ denotes the training and validation corpus of task $T_i$, and $\theta_i$ denotes all training parameters of the dialogue model for $T_i$, which include parameters $\theta^s$/$\theta_i^p$/$\theta^g$ in the shared/private/gating module respectively. we consider a model represented by a parameterized function $f$ with parameters $\theta$.
The model training for all tasks consists of two steps: pre-training and customized model training.

In pre-training, CMAML employs the vanilla MAML to obtain a pre-trained dialogue model as the initial model $\theta$ for all tasks.  
At the beginning of the MAML, $\theta$ are randomly initialized. Then, two main procedures perform iteratively: meta-training and meta-testing. 
In meta-training, MAML first samples a set of tasks ${T_i}{\sim} p(\mathcal{T})$. Then, for each task $i$, MAML adapts $\theta$ to get $\theta_i'$ with the task-specific data, which is, 
 \begin{equation}
\begin{aligned}
\label{euq: meta-training}
\theta_i' = \theta - \alpha \nabla_{\theta}\mathcal{L}_{D_i^{train}}(f(\theta))
\end{aligned}
\end{equation}
In the meta-testing, MAML tests tasks ${T_i}{\sim} p(\mathcal{T})$ with $\theta_i'$ to obtain the losses and then updates $\theta$ by 
\begin{equation}
\begin{aligned}
\label{euq: meta-testing}
\theta = \theta - \beta \nabla_{\theta}\sum_{{T_i}{\sim} p(\mathcal{T})}\mathcal{L}_{D_i^{valid}}(f(\theta_i'))
\end{aligned}
\end{equation}
Here, $\alpha$ and $\beta$ are hyper-parameters.

In standard MAML, each task obtains its parameters $\theta_i$ by fine-tuning the pre-trained $\theta$. 
However, recall that fine-tuning fails to search the best model in the network structure perspective. Also, the generative models are less vulnerable to input changes, thus a few utterances may not be enough to adapt $\theta$ into diverse $\theta_i$ for different tasks. 
To address these issues, we do not perform direct fine-tuning on each task, but design our second training step - Customized Model Training, in which the pre-trained private module can evolve into different structures to capture the characteristics of each task and encourage model diversity. 

\section{Customized Model Training}
After obtaining the pre-trained model $\theta$ from MAML, we employ Customized Mode Training  with the following two updating steps:
\begin{itemize}[wide=0\parindent,noitemsep]
    \item \textbf{Private Network Pruning}. 
    This step is applied for the private module only, which is to differentiate the MLP structure of each task. Each task has a different MLP structure by retaining its own subset of active MLP parameters in order to characterize the uniqueness of this task. 
    \item \textbf{Joint Meta-learning}. 
    In this step, we re-train parameters of all three modules of each task using MAML again, but each private module is with its pruned MLP structure now.
    Also, similar tasks with similar pruned MLP structures are jointly trained in order to enrich the training data.
\end{itemize}
\noindent
In the following, we will describe these two steps respectively as well as the gradient update of the whole dialogue model. 

\subsection{Private Network Pruning}
\label{sec:prune}
After pre-training, dialogue models of different tasks remain the same parameters $\theta$, including $\theta^s$/$\theta^p$/$\theta^g$ in the shared/private/gating module. In this step, the private module with parameters $\theta^p$ will evolve into different structures with parameters $\theta^p_i$ to capture the task's unique characteristics.

First, we fine-tune the whole dialogue model of each task from the MAML initialization with its own training data and add an L-1 regularization on the parameters of the private module.
The goal of L-1 regularization here is to make the parameters sparse such that only parameters beneficial to generate task-specific sentences are active. 

Second, we apply an up-to-bottom strategy to prune the private MLP for each task. 
This is equal to selecting edges in the fully connected layers in the MLP.
We do not prune the layers connected to the input and output of the MLP.
For the rest layers, we start the pruning from the one closest to the output first.
For the $l$-th layer, we consider layers above it ($>l$) are closer to the output, and its lower layers ($<l$) are closer to the input.
When we process the $l$-th layer, its upper layers should already be pruned.
We only keep edges of the current processed layer whose weight excels a certain threshold $\gamma$.
If all edges in the $l$ layer connected to a node is pruned, all edges connected to this node in the $l-1$ layer will also be pruned.
In this way, the parameters in private module $\theta^p$ differentiates into $|T|$ parameters $\theta^p_i$, where each $\theta^p_i$ is a subset of $\theta^p$.
The pruning algorithm described above is illustrated in Algorithm~\ref{alg:pruning_algorithm}. 

\begin{algorithm}[!t]
\IncMargin{2em}
\small
\caption{Private Network Pruning}
\label{alg:pruning_algorithm}
\KwIn{All parameters $\theta^p$ in the private MLP module, the sparsity threshold $\gamma$, the total number of layers $L$ in the private MLP module. 
}
\KwOut{The pruned parameters $\theta^p_i$ in private module for task $T_i$.}
Finetune $\theta^p$ on the training data of $T_i$ with L-1 regularization to otain $\theta_i^p$.\\
\For{$j \in \{1, \ldots, L\}$}
{
    $E_{j} \leftarrow$ All edges (i.e. parameters w.r.t. each edge) in the $j$-th layer in $\theta^p_i$\\
    $N_{j} \leftarrow$ All nodes in the $j$-th layer in $\theta^p_i$\\
} 
$E_{keep} \leftarrow E_{|L|} \cup E_{1}$; $k \leftarrow |L| - 1$;$N_{keep} \leftarrow N_{|L|} \cup N_{1}$.\\
\While{$k > 1$}
{
\For{each edge $e$ in $E_k$}
{
    \If{$e > \gamma$ and the node connected with $e$ in $N_{k+1}$ is in $N_{keep}$}
    {
        $E_{keep} \leftarrow E_{keep} \cup \{e\}$. \\
    }
}
\For{each node $n$ in $N_k$}
{
    \For{each edge $e$ in $E_k$ connected with $n$}
    {
    \If{$e$ in $E_{keep}$}
    {
    $N_{keep} \leftarrow N_{keep} \cup \{n\}$; \\
    break.\\
    }
    }
}
$k \leftarrow k-1$.
}
\Return{$E_{keep}$ as $\theta^p_i$}
\end{algorithm}

\subsection{Joint Meta-learning} 
So far, every task has a unique network structure in its private module. 
Now we jointly train the whole dialogue models of all tasks. 

We start from the pre-trained MAML initialization again.
For the shared and gating modules, all tasks share the same parameters, and they are trained with all training data. 
The private module, which is to capture the uniqueness of each task, is supposed to be trained on task-specific data. However, we do not have sufficient training data for each task in the few-shot setting, thus the private module may not be trained well. 
Fortunately, all private modules evolve from the same MLP structure, and similar tasks naturally share overlapped network structures, i.e. remaining edges after pruning are overlapped. 
This inspires us to train each edge in the private MLP by all training samples of tasks in which this edge is not pruned. 

Concretely, we train the private MLP in this way: 
for each edge $e$ in the MLP, if it is active in more than one tasks, its corresponding parameters $\theta_{e}^p$ are updated on the data of all task $j$'s, 
in which the edge is active, i.e. $\theta_e^p \in \theta_j^p$: 
\begin{equation}
\begin{aligned}
\label{euq:fusion_train}
\theta'^p_{e} = \theta^p_{e} - \alpha \nabla_{\theta^p_{e}}\sum_{T_j: \theta_e^p \in \theta^p_j}\mathcal{L}_{D_j^{train}}(f(\theta^p_j))\\
\end{aligned}
\end{equation}
where each $\theta^p_i$/$\theta'^p_i$ only contains the $\theta^p_e$/$\theta'^p_e$'s of all active edges in the $i$-th task. 

During meta-testing, the loss is accumulated by the tasks that use the corresponding dialogue models, so $\theta^p$ is updated as,
\begin{equation}
\begin{aligned}
\label{euq:fusion_test}
\theta^p= \theta^p - \beta \sum_{{T_i}{\sim}  p(\mathcal{T})} \nabla_{\theta^p_i} \mathcal{L}_{D_i^{valid}}(f(\theta'^p_i))
\end{aligned}
\end{equation}

\subsection{Gradient Updates}

We summarize the gradient updates of the three modules in our proposed dialogue model during customized model training in Algorithm~\ref{alg:Customized_alg}.
For the shared and gating module, gradients are updated in the same way as MAML.
The update of the private module is replaced by the above Eq.~\ref{euq:fusion_train} and Eq.~\ref{euq:fusion_test} introduced in joint meta-learning.

The loss function used to calculate the gradients in our model is the negative log-likelihood of generating the response $r$ given the input query $q$ as,
\begin{equation}
\begin{aligned}
\label{euq: gating}
\mathcal{L} = -  \log p(r|q, \theta^s,  \theta^p, \theta^g)
\end{aligned}
\end{equation}

\begin{algorithm}[!t]
\IncMargin{2em}
\small
\caption{Customized Model Training}
\KwIn{ The distribution over the task set $p(\mathcal{T})$, the step size $\alpha$ and $\beta$.
}
\KwOut{The customized dialogue models $\mathcal{\theta}^s \cup {\theta}^p_i \cup {\theta}^g $ for every task $T_i$.}

\For{each $T_i$ in $T$}
{
$\theta^p_i$ $\leftarrow$  \mbox{Private\_Network\_Pruning}($T_i$). \\
}
\While{not converge}
{
Sample a batch of tasks ${T_i}{\sim} p(\mathcal{T})$. \\
\For{each sampled task $T_i$}
{
Adapt $\theta^s/\theta^g$ to $\theta'^s_i/\theta'^g_i$ with $D_i^{train}$ using Eq.~\ref{euq: meta-training};\\
Adapt $\theta^p_i$ to $\theta'^p_i$   with $D_i^{train}$ using Eq.~\ref{euq:fusion_train}.
}
Update $\theta^s,\theta^g$ with $D_i^{valid}$ using Eq.~\ref{euq: meta-testing}. \\
Update $\theta^p_i$ with $D_i^{valid}$ using Eq.~\ref{euq:fusion_test}.
}
\Return{$\mathcal{\theta}^s \cup {\theta}^p_i \cup {\theta}^g $}
\label{alg:Customized_alg}
\end{algorithm}

\section{Experiments}

\subsection{Datasets}
We perform experiments in Persona-chat~\cite{paml} and MojiTalk~\cite{mojitalk}, which are treated as few-shot dialogue generation tasks in previous work~\cite{persona,paml,mojitalk,emotion-chat}. 
Persona-chat has 1137/99/100 users for training/validation/evaluation, and each user has 121 utterances on average. We follow the previous work~\cite{paml} and concatenate all the contextual utterances including the query as the input sequence. We regard building a dialogue model for a user as a task on this dataset. 
MojiTalk has 50/6/8 emojis for training/validation/evaluation. Each training/validation emoji has 1000 training samples on average, and each evaluation emoji has 155 samples on average. 
We regard generating responses with a designated emoji as a task.
On both datasets, the data ratio for meta-training and meta-testing is 10:1.


\subsection{Implementation Details}
We implement our shared module based on the Seq2seq model with pre-trained Glove embedding~\cite{glove} and LSTM unit, and use a 4-layer MLP for the private module\footnote{Code is available at https://github.com/zequnl/CMAML}. The dimension of word embedding, hidden state, and MLP's output are set to 300. In CMAML, we pre-train the model for 10 epochs and re-train each model for 5 steps to prune the private network. The L-1 weight in the re-training stage is 0.001, and the threshold $\gamma$ is 0.05. We follow other hyperparameter settings in \citet{paml}.

\subsection{Competing Methods}
\begin{itemize}[wide=0\parindent,noitemsep]
\item \textbf{Pretrain-Only}: We pre-train a unified dialogue generation model with data from all training tasks then directly test it on the testing tasks. We try three base generation models: the \textit{Seq2seq}~\cite{seq2seq-attention} and the \textit{Speaker} model~\cite{lijiwei} and the \textit{Seq2SPG} proposed in Section\ref{sec:dialogue model}. \textit{Speaker} incorporates the task (user/emoji) embeddings in the LSTM cell, and the task embeddings of testing tasks are random parameters in this setting.

\item \textbf{Finetune}: We fine-tune the pre-trained models on each testing task, denoted as \textit{Seq2seq-F}, \textit{Speaker-F} and \textit{Seq2SPG-F}.

\item \textbf{MAML}~\cite{paml}: We apply the MAML algorithm on the base model Seq2seq and Seq2SPG, and note them as \textit{MAML-Seq2seq} and \textit{MAML-Seq2SPG}. MAML-Seq2SPG uses the same base model as the proposed CMAML but does not apply the pruning algorithm, which helps to verify the effectiveness of the pruning algorithm and joint meta-learning. Note that We did not apply MAML on Speaker model as it shows no improvement comparing with Seq2seq. 

\item \textbf{CMAML}: We try two variants of our proposed algorithm. \textit{CMAML-Seq2SPG} is our full model (equal to CMAML in previous sections), where the dialogue Seq2SPG is the base model and pruning algorithm is applied for customizing unique model structures for tasks. 
\textit{CMAML-Seq2SP$'$G} uses a different base model noted as Seq2SP$'$G, where the private module
only takes the output of the shared module as the input. Pruning algorithm is also applied in private module for network customization. 
\end{itemize}

\begin{table*}[!t]
\scriptsize
\centering
\begin{tabular}{lcc|cccc|ccccc}
\hline
\textbf{\multirow{2}{*}{Method}} & \multicolumn{2}{c|}{\textbf{Human Evaluation}} & \multicolumn{4}{c|}{\textbf{Automatic Metrics}} &
\multicolumn{2}{c}{\textbf{Model Difference}} \\
\cline{2-3}\cline{4-7}\cline{8-9}
& \textbf{Quality} & \textbf{Task Consistency} & \textbf{PPL}  & \textbf{BLEU} & \textbf{Dist-1}  & \textbf{C score/E-acc}  & \textbf{Diff Score} & \textbf{$\Delta$ Score}\\
\hline
\textbf{Persona-Chat} &&&&&&&&\\
Seq2seq  & 0.67 & 0.10 & 37.91 &	1.27 & 0.0019	 &-0.16	   & 0.00 &0.00\\
Speaker  &0.85& 0.10 &40.17  &1.25	& 0.0037	 &-0.14	  & 0.00 & 0.00\\
Seq2SPG & 0.67 &0.03 & 36.46 &	1.41& 0.0023	 &	-0.14  & 0.00 & 0.00\\
\hline
Seq2seq-F & 0.78 & 0.11   &33.65  & 1.56 & 0.0046& 	-0.05		 &17.97  & 9.19\\
Speaker-F & 0.87 & 0.25  & 35.61 &1.52 &0.0059	 &	0.03	 &  285.11& 143.90\\
Seq2SPG-F & 0.7 & 0.07  & \textbf{32.68} &1.54 &	0.0045 &	-0.05	 & 292.85 & 156.30\\
\hline
MAML-Seq2seq & 0.97 & 0.37   & 37.43 &1.54 &0.0087  &0.14	 & 134.01 & 67.79\\
MAML-Seq2SPG  & 0.85  &	0.36  &	 35.89 & \textbf{1.70}  &0.0074	  & 0.16 & 401.28& 198.90\\
\hline
CMAML-Seq2SP$'$G  & 0.98 & 0.58 & 37.32 & 	1.43  &0.0089 &	0.15	 & 479.21 & 238.64\\
CMAML-Seq2SPG & \textbf{1.15} & \textbf{0.69}  & 36.30 &\textbf{1.70}  &\textbf{0.0097} &		\textbf{0.18} &\textbf{514.44} &\textbf{263.82}\\
\hline
\hline
\textbf{MojiTalk} &&&&&&&&\\
Seq2seq & 0.56  & 0.39 &  218.95 & 0.36   & 0.0342 &0.73	  & 0.00 & 0.00\\
Speaker & 0.38  & 0.26 & 418.96  &  0.19&\textbf{0.0530} &0.70	  & 0.00 & 0.00\\
Seq2SPG  & 0.77  &0.46  & 158.74  &  0.64&0.0239 &	0.74  & 0.00 & 0.00\\
\hline
Seq2seq-F & 0.50  &0.35  & 217.60  & 0.40 &0.0326	  &0.72	 & 15.96&8.88  \\
Speaker-F &  0.39 & 0.25 & 403.92  & 0.21  &0.0528 &0.72	   &39.08& 29.10\\
 Seq2SPG-F &  0.76 & 0.47 & \textbf{157.92} & 0.65  &0.0228 &	0.74   & 72.43&40.94\\
\hline
MAML-Seq2seq & 0.66  &0.29 & 179.02 & 0.54 & 0.0109 & 0.70	  &183.05& 117.09 \\
MAML-Seq2SPG & 0.71  &0.40	  &	181.56  &	0.73 & 0.0246  &0.74  &306.40& 176.31\\
\hline
CMAML-Seq2SP$'$G &0.64 & 0.32 & 172.92  & 0.76 &	0.0102  &0.75	&  142.90 &81.15 \\
CMAML-Seq2SPG &  \textbf{0.78} & \textbf{0.49} & 185.97  &  \textbf{0.85}&	 0.0210 &\textbf{0.77}	  &  \textbf{345.42}&\textbf{190.64}\\
\hline
\end{tabular}
\caption{Overall performance in Persona-chat (top) and MojiTalk (bottom) dataset in terms of quality (Human, Perplexity, BLEU), diversity (Dist-1), task consistency (Human, C score, E-acc), structure differences among tasks (Diff Score ($\times10^{-10}$)), model change after adaptation ($\Delta$ score ($\times10^{-10}$)).}
\label{tab:overall}
\end{table*}

\subsection{Evaluation Metrics}
{\bf Automatic Evaluation.} We performed automatic evaluation metrics in three perspectives:
\begin{itemize}[wide=0\parindent,noitemsep]
    \item Response quality/diversity:  We use \textbf{BLEU}~\cite{bleu} to measure the word overlap between the reference and the generated sentence;  \textbf{PPL}, the negative logarithm of the generated sentence; \textbf{Dist-1}~\cite{lijiwei-diversity,ijcnlp,dpp} to evaluate the response diversity, which calculates the ratio of distinct 1-gram in all test generated responses. 
    
    \item Task consistency: We use \textbf{C score}~\cite{paml} in Persona-chat, which uses a pre-trained natural language inference model to measure the response consistency with persona description, and \textbf{E-acc}~\cite{mojitalk} in MojiTalk, which uses an emotion classifier to predict the correlation between a response and the designated emotion.

    \item Model difference: It is hard to measure the model’s ability of customization as we do not have the ground-truth model. Hence, we define the average model difference of pairwise tasks as the \textbf{Diff Score} of each method, and the model difference of a method before and after fine-tuning as \textbf{$\Delta$ Score}. 
    The model difference between $T_i$ and $T_j$ is the Euclidean distance of their parameters normalized by their parameter count:
    $D(T_i, T_j) = \frac{||\theta_i-\theta_j||^2}{M}$.
    Here, $\theta_i$/$\theta_j$ includes all model parameters of this task, $M$ is the total parameter number of the model.
    A set of models that capture the unique characteristics of each task should be different from each other and will have a higher Diff score, indicating that a large Diff score is a sufficient condition for a strong customization ability. Similarly, a model that changes a lot for task specific adaptation during fine-tuning will achieve a higher \textbf{$\Delta$ Score}, indicating that \textbf{$\Delta$ Score} is also a sufficient condition for a good adaptation ability.

\end{itemize}

\noindent 
\textbf{Human Evaluation.} 
We invited 3 well-educated graduated students to annotate the 100 generated replies for each method. 
For each dataset,
the annotators are requested to grade each response in terms of ``quality" and ``task consistency" (i.e. personality consistency in Persona-Chat and emoji consistency in MojiTalk) independently in three scales: 2 (for good), 1 (for fair) and 0 (for bad).
``quality" measures the appropriateness of replies, and we refer 2 for fluent, highly consistent (between query and reply), and informativeness, 1 for few grammar mistakes, moderate consistent, and universal reply, and 0 for incomprehensible or unrelated topic. ``task consistency" measures whether a reply is consistent with the characteristics of a certain task, and we refer 2 for highly consistent, 1 for no conflicted and 0 for contradicted. Notice that the user description (Persona dataset) and sentences with a certain emoji (Mojitalk dataset) are provided as the references. 
Volunteers, instead of authors, conduct the double-blind annotations on shuffled samples to avoid subjective bias.

\subsection{Overall Performance}

\begin{table*}[!t]
\scriptsize
\centering
\begin{tabular}{lccc|ccc||ccc|ccc}
\hline
\textbf{\multirow{2}{*}{Method}} & \multicolumn{3}{c|}{\textbf{100-shot}} & \multicolumn{3}{c||}{\textbf{110-shot}} &
\multicolumn{3}{c|}{\textbf{Similar Users}} & \multicolumn{3}{c}{\textbf{Dissimilar Users}}
\\
\cline{2-4}\cline{5-7}\cline{8-10}\cline{11-13}

& \textbf{PPL}  & \textbf{BLEU} & \textbf{C score} & \textbf{PPL}  & \textbf{BLEU} & \textbf{C score} 
& \textbf{PPL}  & \textbf{BLEU} & \textbf{C score} & \textbf{PPL}  & \textbf{BLEU} & \textbf{C score}
\\
\hline
Seq2seq & 38.13  & 1.19  &-0.11	 & 37.58 & 1.29& -0.15& 76.54 &	 1.49 &  -0.03& 42.87 &  1.10 &	 -0.10  \\
Speaker & 40.95 & 1.02 &-0.25 &  42.59 &	1.27 & -0.06  & 162.44	& 0.65 & -0.09 & 46.86 & 1.11  &-0.13	 \\
Seq2SPG & 39.75 & 1.27 & -0.10& 37.71  &1.30 &  -0.15 & 73.58	& 1.32 & -0.04&42.21 &1.14   &-0.22\\
\hline
Seq2seq-F & \textbf{34.86} & 1.39  &	-0.03& \textbf{34.14}& 1.52&-0.10 &  74.53 	& 1.53& -0.07 & 42.33  &1.33   &-0.06	 \\
Speaker-F &37.11	 & 1.30 & -0.16  & 39.10  &  1.36	&-0.06
& 103.81	& 1.04 & \textbf{0.04} & 40.47 & 1.40  &0.01	\\
Seq2SPG-F &	37.19 & 1.31 & 0.00  & 37.00  &  1.33	&-0.15 & \textbf{70.15}	& 1.44 & -0.04 & \textbf{36.22} & 1.35 & -0.05	\\
\hline
MAML-Seq2seq &  36.94& 1.47& 0.03& 37.20 &  1.53 &0.07	 & 83.17	& 1.52 & -0.08 & 39.67 &  1.34 &0.06	 \\
MAML-Seq2SPG & 36.50 &	 1.52 & 0.11 & 35.98  & 1.47 & 0.13
 & 82.37 &	1.52 & -0.06 & 39.41 & 1.41 &  0.12\\
 \hline
CMAML-Seq2SP$'$G & 37.18 &	1.46 & 0.11 &37.08 & 1.44 & 0.09
& 82.56 &	1.50 &	0.00 & 40.50	&1.40  &0.13	\\
CMAML-Seq2SPG & 36.52 & \textbf{1.52}	 & \textbf{0.14}	 & 36.44 & \textbf{1.57} & \textbf{0.15}
& 82.78 &	\textbf{1.56} & -0.07 & 39.55	& \textbf{1.43} &\textbf{0.16}	\\
\hline
\end{tabular}
\caption{The performance on the Persona-chat dataset for impact factor analysis. The left figure is about the few-shot settings and the right is about the task similarity.}
\label{tab:imapact_factor}
\end{table*}

\noindent\textbf{Quality/Diversity.}
In the Persona-chat dataset, \textbf{Pretrain-Only} methods provide the borderlines of all methods. In \textbf{Pretrain-Only}, \textit{Seq2SPG} achieves the best performance in terms of both automatic and human measurements, indicating the appropriateness of the proposed model structure. \textbf{Finetune} methods are better than \textbf{Pretrain-Only} methods in most cases. \textbf{MAML} methods have no better performance on BLEU scores than \textbf{Finetune} methods but have relatively higher Dist-1 scores. This indicates that MAML helps to boost response diversity. Enhanced with the proposed pruning algorithm, we can see great improvement for \textbf{CMAML} methods against all the competing methods on both quality and diversity measurements. Particularly, our full model \textit{CMAML-Seq2SPG} shows clearly better performance and the reasons can be ascribed to two aspects: firstly, the proposed \textit{Seq2SPG} has a better model structure for our task and secondly, the pruning algorithm makes the models more likely to generate a user-coherent response. 

Most of the performance of the competing methods in the MojiTalk dataset is similar to the Persona-chat dataset, while one difference is that \textit{Speaker} achieves the highest Dist-1 score among all the methods. By carefully analyzing the generated cases, we find all non-meta-learning methods (\textbf{Pretrain-Only} and \textbf{Finetune}) consistently produce random word sequences, which means they completely fail in the few-shot setting on this task. However, meta-learning-based methods survive.

\noindent\textbf{Task Consistency.} On both datasets, \textbf{Finetune} methods make no significant differences on C score, E-acc and Task Consistency when compared with \textbf{Pretrain-Only} methods, which means that simple fine-tuning is useless for improving the task consistency. All meta-learning methods including \textbf{MAML} and \textbf{CMAML} outperforms \textbf{Finetune}.
Compared with \textit{MAML-Seq2seq} and \textit{MAML-Seq2SPG}, \textit{CMAML-Seq2SPG} obtain 22.2\%/12.5\% and 11.8\%/5.6\% improvement on C score and E-acc. It means that the private modules in \textit{CMAML-Seq2SPG} are well pruned to better well describes the unique characteristics of each task. 

We also observe that in MojiTalk, \textit{CMAML-Seq2SPG} achieves good improvement compared with other baselines on the BLEU score but a limited improvement on E-acc and task consistency score when compared with Persona-chat. This tells that when the training data is limited, the generative models tend to focus on the correctness of the response rather than the task consistency. 

By jointly analyzing the response quality and task consistency measurement, we can easily draw the conclusion that the responses produced by our algorithm in \textit{CMAML-Seq2SPG} not only is superior in response quality but also caters to the characteristics of the corresponding task.

\noindent\textbf{Model Differences.}
Even though a high difference score among tasks does not indicate each model has captured its unique characteristics, a set of models that can capture the characteristics of themselves will have a higher different score. Hence, we present the difference scores of competing methods as a reference index. 
In Table \ref{tab:overall}, we can see that fine-tuning on non-meta-learning methods (\textbf{Pretrain-Only} and \textbf{Finetune}) does not boost the model differences between tasks. MAML helps to increase the model differences but is not as good as the proposed \textbf{CMAML} methods. \textit{CMAML-Seq2SPG} achieves the highest model difference scores on two datasets as it distinguishes different tasks in both parameter and model structure level.

A higher $\Delta$ score of a method means its produced dialogue models are more easy to fine-tune. All non-meta-learning methods have so much lower $\Delta$ scores than \textbf{MAML} methods. \textit{CMAML-Seq2SPG} has the highest scores on both datasets, indicating that the active edges in the private module are more likely to be fine-tuned to better fit the corpus of the corresponding tasks. We also observe that \textit{CMAML-Seq2SP$'$G} has relatively low $\Delta$ scores, which indicates its base generation model \textit{Seq2S$'$G} is not as good as \textit{Seq2SPG}.

\subsection{Impact Factors}
We further examine two factors that may have a great impact on the performance: the quantity of training data and the similarity among tasks.

\noindent\textbf{Few-shot Settings.}
We only use Persona-chat dataset for analysis, because MojiTalk has too little data to further decrease.  
In Persona-chat, each user has 121 training samples on average, and we evaluate all the methods in a 100 and 110 samples setting (both in train and test) in Table \ref{tab:imapact_factor} because all the methods tend to produce random sequences when each task contains less than 100 samples.

For non-meta-learning methods including \textbf{Pretrain-Only} and \textbf{Finetune}, the quality scores improve as the quantity of training data increases, while the C scores almost remain the same as these methods are not sensitive to the differences among tasks. \textbf{MAML} methods have not changed too much on BLEU scores along with the data growth, but its C scores keep increasing. Both the BLEU score and C score of \textit{CMAML-Seq2SPG} keep increasing with the data growth, and it always achieves the best performance among all the tasks. This proves that the customized generative models are suitable for the corresponding tasks and can always take the full potential of the training data.


\noindent\textbf{Task Similarity.}
Again, we only use the Persona-chat dataset because we cannot define similarities among emojis. We construct two datasets: one contains 100 similar users and another contains 100 dissimilar users (both in train and test).

The performance of all the methods is close to each other in the similar-user setting. It means meta-learning-based methods have no advantage for similar tasks. In the dissimilar-users setting, \textit{CMAML-Seq2SPG} performs best on the C score and BLEU. We draw a conclusion that user similarity influences the performance of our model. Compared to that in dissimilar-users setting, the BLEU in the similar-users setting is high, but the C score is low. The possible reason is that generative models do not distinguish similar tasks and regard all tasks as one task in training. 

\begin{table}[!t]
\scriptsize
\centering
\begin{tabular}{l|l}
\hline
\multirow{3}{*}{Persona} 
 & I also love vintage cars.  \\
 & I am a pediatrician.  \\
 & I love running and reading.  \\
\hline\hline 
{Query} &  Singing karaoke is a talent of mine. Do you sing too? \\
{Response} & Not really. I am into running, books and old cars. \\
\hline \hline 
{Seq2seq} &    I do not have any pets. I do not have any pets. \\
{Speaker}        & No, I do not. I do not have any. \\
{Seq2SPG} & No , I do not have any pets.\\
\hline
{Seq2seq-F} &  I do not have any pets. I do not have any pets.   \\
{Speaker-F}    & No, I do not. I do not have any. \\
{Seq2SPG-F} & No , I do not have any pets.\\
\hline 
\tiny{MAML-Seq2seq}    & Yes I do. I am a nurse. \\
\tiny{MAML-Seq2SPG}    & I like to listen to jazz and jazz .\\
\hline 
  \tiny{CMAML-Seq2SP$'$G}   & Yes, I am a doctor. I am a pediatrician.   \\
    \tiny{CMAML-Seq2SPG}     & Yes, I am a pediatrician. What do you do for a living?  \\
\hline
\end{tabular}
\caption{A case in Persona-chat dataset.}\label{tab:case}
\end{table}

\subsection{Case Study}
We only present one case in the Persona-chat dataset due to the limited space in Table~\ref{tab:case}. \textbf{Pretrain-Only} and \textbf{Finetune} methods produce general responses with less information. \textbf{MAML} methods tend to generate diverse responses as their initial parameters are easier to be finetuned. Even though the user profiles are not used for training, \textit{CMAML-Seq2SPG} can quickly learn the persona information ``pediatrician" from its training dialogues while other baselines can not. From another perspective, the pruned private module in \textit{CMAML-Seq2SPG} can be regarded as a special memory that stores the task-specific information without explicit definition of memory cells.

\section{Conclusion}
In this paper, we address the problem of the few-shot dialogue generation. We propose CMAML, which is able to customize unique dialogue models for different tasks.
CMAML introduces a private network for each task's dialogue model, whose structure will evolve during the training to better fit the characteristics of this task. The private module will only be trained on the corpora of the corresponding task and its similar tasks. 
The experiment results show that CMAML achieves the best performance in terms of response quality, diversity and task consistency. We also measure the model differences among tasks, and the results prove that CMAML produces diverse dialogue models for different tasks.

\section{Acknowledgement}
This paper is partially supported by National Key Research and Development Program of China with Grant No. 2018AAA0101900/2018AAA0101902, Beijing Municipal Commission of Science and Technology under Grant No. Z181100008918005, and the National Natural Science Foundation of China (NSFC Grant No. 61772039, No. 91646202, and No. 61876196).

\bibliographystyle{named}
\bibliography{acl2020}

\end{document}